\pdfoutput=1
\documentclass[11pt]{article}
\usepackage{booktabs}
\usepackage{array}
\usepackage{graphicx}

\usepackage[]{acl}
\usepackage{newtxmath}

\usepackage{times}
\usepackage{latexsym}
\usepackage{url}
\usepackage{multirow}
\usepackage{subcaption}

\usepackage[T1]{fontenc}

\usepackage[utf8]{inputenc}

\usepackage{microtype}

\title{Unsupervised Reinforcement Adaptation for Class-Imbalanced Text Classification}

\usepackage{algorithm}
\usepackage{algorithmic}

\author{Yuexin Wu \\
  University of Memphis\\
  \texttt{ywu10@memphis.edu} \\\And
  Xiaolei Huang \\
  University of Memphis  \\
  \texttt{xiaolei.huang@memphis.edu} \\}

\begin{document}
\maketitle
\begin{abstract}

Class-imbalance naturally exists when train and test models in different domains.
Unsupervised domain adaptation (UDA) augment model performance with only accessible annotations from the source domain and unlabeled data from the target domain.
However, existing state-of-the-art UDA models learn domain-invariant representations and evaluate primarily on class-balanced data across domains.
In this work, we propose an unsupervised domain adaptation approach via reinforcement learning that jointly leverages feature variants and imbalanced labels across domains.
We experiment with the text classification task for its easily accessible datasets and compare the proposed method with five baselines.
Experiments on three datasets prove that our proposed method can effectively learn robust domain-invariant representations and successfully adapt text classifiers on imbalanced classes over domains.
The code is available at \url{https://github.com/woqingdoua/ImbalanceClass}

\end{abstract}

\section{Introduction}




\textit{Unsupervised domain adaptation} (UDA) is to find a shared feature space that is predictive across target and source domains~\cite{ramponi2020neural}.
The shared space, \textit{domain-independent} feature set, allows transferring of trained text classifiers from the source domain to the target domain.
Methods to find the space have two major directions, pivot feature~\cite{blitzer-etal-2006-domain, daume2007frustratingly, ziser-reichart-2018-pivot, ben-david2020perl} and adversarial learning~\cite{ganin2015unsupervised, chen2020adversarial, du-etal-2020-adversarial}.
The pivot-based method selects a subset of shared features, called pivots, which learn important cross-domain information to represent shared feature space.
Adversarial learning approaches the shared feature space by reducing document features' capability to distinguish source and target domains.
The common method to achieve this is Gradient Reversal Layer (GRL)~\cite{ganin2015unsupervised} aiming to reduce domain-specific patterns.
However, the UDA approaches primarily focus on feature shifts ($p(X_{source}) != P(X_{target})$) while ignore possible class shifts ($p(Y_{source}) != p(Y_{target})$) across domains.

\textit{Class-imbalance} naturally exists in data when label distributions across domains~\cite{cui2017effect, cheng2020representation} are different.
Under the class-imbalanced scenario, the label distribution is imbalanced across domains, and the label distributions in source and target domains are not the same.
Given the widely used Amazon data~\cite{ni2019justifying} as an example, the Book reviews may have more positive reviews than negative reviews, and the Kitchen may have a lower ratio of negative reviews.
However, evaluating unsupervised domain adaptation under the class-imbalanced scenario is under-examined than the ideal scenario of the class-balanced benchmark.
A wide evaluation benchmark of UDA for text classifiers is extracted from the Amazon review~\cite{blitzer-etal-2006-domain}.
The data has the same balanced-class distributions for both source and target domains.
Such a well-balanced label distribution may make existing UDA models inapplicable to the real-world scenario, where class distributions can shift across domains.

In this study, we proposed an \textit{unsupervised reinforcement adaptation model} (URAM) for text classifiers under the UDA setting that only labeled source data and unlabeled target data are available.
Specifically, we propose a neural mask mechanism to generate domain-dependent and -independent feature representations and a reward policy using a critic value network~\cite{NIPS1999_6449f44a} (CRN) to learn optimal domain-independent representations.
The reward policy optimizes the URAM via three joint reward factors, label, domain, and domain distance.
While the label reward aims to encourage text classification models on domain-independent features to predict correct document classes, the domain and domain distance rewards reduce domain variations of domain-dependent feature representations between source and target domains.
We compare our reinforcement adaptation model with five baselines and experiment on four class-imbalanced data with both binary and non-binary labels.
The results using the F1-score demonstrate the effectiveness of our reinforcement learning model that outperforms the baselines by 3.13 on average.
The main contributions of this paper are as follows:
\begin{itemize}
    \item We propose a reinforcement learning model for unsupervised domain adaptation that jointly leverages cross-domain variations and classification performance.
    \item We experiment UDA approaches on the class-imbalanced scenario that label distributions are different across domains. The class-imbalanced scenario is under-explored among  the UDA models .
    \item We conduct an extensive ablation analysis that demonstrates how the reinforcement model can coherently combines both pivot and adversarial directions of unsupervised domain adaptation.
\end{itemize}


\section{Background}
This section briefly recaps the concepts of unsupervised domain adaptation (UDA) and reinforcement learning.

\subsection{UDA for Class-Imbalanced Data}
UDA assumes a labeled dataset with $\mathcal{D}_{S}=\left\{\left(x_{s}^{i}, y_{s}^{i}\right)\right\}_{i=1}^{n_{s}}$ from source domain and a unlabeled data $\mathcal{D}_{T}=\left\{x_{t}^{j}\right\}_{j=1}^{n_{t}}$ from target domain, data distributions of the two domains are different, $p(x_{s}) \neq p(x_{t})$, and the two domains share the same number of \textit{unique} annotations.
UDA is to find a common feature space aligning source and target domains so that $f(p(x_{s})) \approx p(x_{t})$
However, class-imbalanced data naturally exist in UDA tasks that may cause inefficient knowledge transfer~\cite{ramponi2020neural}.
We assume both data and labels are not equally distributed in this work.

\subsection{Reinforcement Learning}
Actor-Critic~\cite{NIPS1999_6449f44a} is an reinforcement learning algorithm that combines Actor and Critic networks. 
Critic, a value network (denote as $V_{\theta_{c}}$), estimates rewards at state $s_{t}$ and is optimized by state difference error as follows
\begin{equation}
    \mathcal{L}(\theta _{c})=\left\|{V}_{\theta _{c}}\left({s_{t}}\right)-r\left(\mathbf{s}_{t}, {a}_{t}\right) - {V}_{\theta_{c}}\left(\mathbf{s}_{t+1}\right)\right\|^{2}
\end{equation}
where $r(s_{t},a_{t})$ is a target reward and tells us the reward for taking action $a$ in state $s$.
The actor is a policy function that gives us the probability of taking action $a$ in the state $s$.
The actor decides which action should be taken, and the critic evaluates how good the action is and how it should adjust. 
The learning of the actor ($\theta_{a}$) is based on policy gradient approach as the following
\begin{equation}\label{op1}
    \mathcal{L}^{A}(\theta_{a}) = \sum_{t} \log \pi_{\theta_{a}}\left(a_{t}, s_{t}\right) A\left(s_{t}, a_{t}\right)
\end{equation}
, where $A\left(s_{t}, a_{t}\right)=r\left(s_{t}, a_{t}\right)+\gamma V\left(s_{t+1}\right)-V\left(s_{t}\right)$. 
$\gamma$ is a decay factor that discounts rewards backward over steps.
To encourage the actor to explore more actions, the algorithm adds an entropy penalty,
\begin{equation}\label{op2}
    \mathcal{L}^{S}(\theta_{a})= - \sum_{a} \pi_{\theta}(a \mid s) \log \pi_{\theta_{a}}(a \mid s)
\end{equation}
The overall objective is as following,
\begin{equation}
    \mathcal{L} = \min \{\mathcal{L}(\theta_{c}) - (\mathcal{L}^{A}(\theta_{a}) + \mathcal{L}^{s}(\theta_{a})) \}
\end{equation}

\begin{figure*}[ht!]
\centering 
\includegraphics[width=0.98\textwidth]{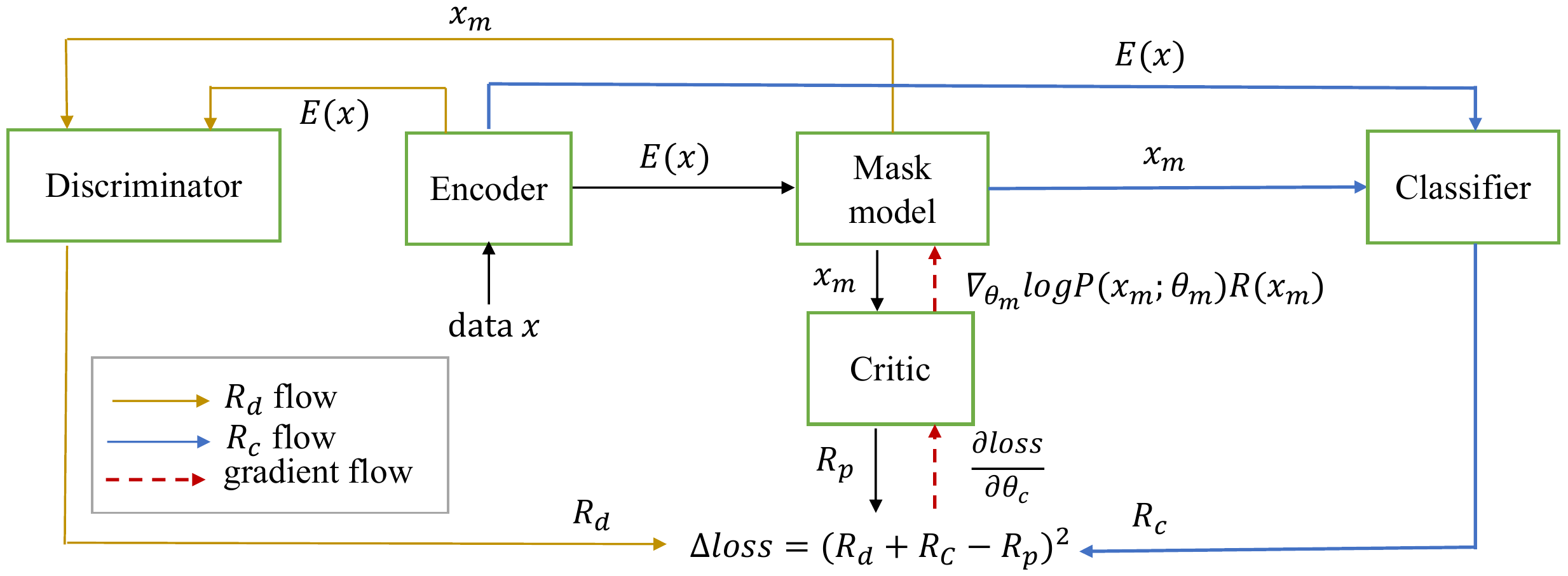}
\caption{Illustration of URAM learning process. The yellow route is the $R_{d}$ prediction progress, measuring by the confusion of the discriminator for $x_{m}$ and $E(X)$. The blue route calculates the $R_{d}$ by the consistency of the classifier for $x_{m}$ and $E(X)$. The overall reward is made by  $R_{d}$ and $R_{c}$.  The critic updates parameters by minimizing the mean square between the ground truth reward ($R_{d} + R_{c}$) and the predictive reward ($R_{c}$). The mask  model learns from the policy gradient.}
\label{mr}
\end{figure*}
\section{Unsupervised Reinforcement Adaptation Model}

In this section, we present details of the Unsupervised Reinforcement Adaptation Model (URAM) in Figure~\ref{mr}.
The URAM trains classifiers on the labeled data from the source domain and unlabeled data from the target domain.
The model contains three major modules: 1) a base model; 2) adversarial learning; 3) reinforcement learning.
%
 
\subsection{Based Model}\label{in-domin training}
Our based model consists of an encoder and a classifier.
The encoder extracts features from input documents, and the classifier predicts document labels.
The based model takes a regular in-domain training method with $n_{s}$ labeled samples from the source domain
 \begin{equation}\label{step a}
     \min_{\theta_{e},\theta_{cla}} \sum_{i}^{n_s} (\mathcal{L}(C(E(x_{s}^{i},\theta_{e}),\theta_{cla}),y_{s}^{i})
 \end{equation}
, where $\theta_{e}, \theta_{cla}$ are the parameters of the encoder and classifier respectively. $\mathcal{L}(\cdot)$ is the cross-entropy loss.

\subsection{Adversarial Learning}
We propose an adversarial learning using a mask strategy to learn domain-independent representations, which are transferable across domains.
Domain adversarial learning~\cite{ganin2015unsupervised} learns domain-independent features by reduce domain predictability of text classifiers, which is a common adversarial strategy in the UDA~\cite{ramponi2020neural}. 
The domain adversarial learning deploys a discriminator ($\theta_d$) to predict domains by minimizing the classification error:
\begin{equation}\label{discriminator}
    \min_{\theta_{d}} (\mathcal{L}(D(E(X_{s}),\theta_{d}),\vmathbb{1})+\mathcal{L}(D(E(X_{t}),\theta_{d}),\vmathbb{0})).
\end{equation}
However, the uncertainty is a major issue that can lead to uncontrollable learning process~\cite{long2018conditional, ramponi2020neural} and easily fail to yield domain-independent representations.


Therefore, we propose a \textit{mask model} to extract domain-independent features.
Intuitively, if the discriminator uses the generated features from the mask model and fails to recognize domains of input data, then this indicates the features generated by the mask model are domain-independent. 
Therefore, our first goal is to maximize the loss of the discriminator as the following formulation:
\begin{equation}\label{rd}
\begin{aligned}
    R_{d} = \max_{\theta_{m}} (\mathcal{L}(D(x_{m}^{s},\theta_{m}),\theta_{d}),\vmathbb{1})+\\
    \mathcal{L}(D(x_{m}^{t},\theta_{m}),\theta_{d}),\vmathbb{0}))
\end{aligned}
\end{equation}
where $x_{m}^{s} = M(E(X_{s})) * E(X_{s})$ and $x_{m}^{t} = M(E(X_{t})) * E(X_{t})$.
The mask model generates domain-independent representations by learning how to transform domain-dependent features and capture common knowledge cross domains.

\begin{figure*}[ht!]
\centering 
\includegraphics[width=0.95\textwidth]{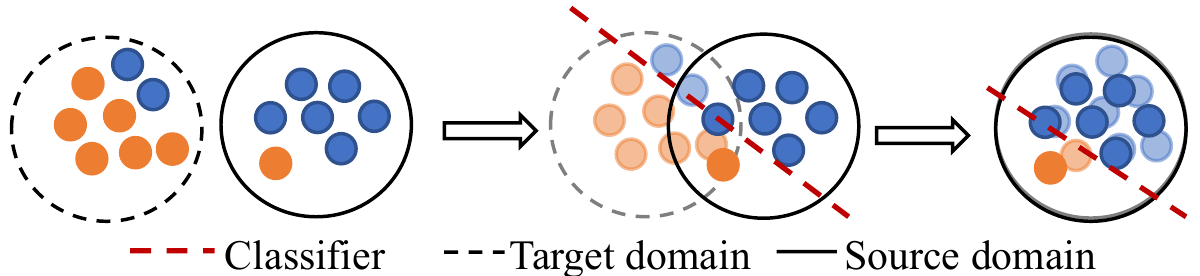}
\caption{Illustration of alignment process for the class-imbalanced data.}
\label{pp}
\end{figure*}

The second objective of the mask model is for class-imbalanced distributions between source and target domains.
Misclassifications occur when there is a class distribution discrepancy between training and test domains.
The optimal domain adaptation is in the second stage of Fig.~\ref{pp}, however the class-imbalance may lead to misalignment in the third stage.
As shown in the Fig.~\ref{pp}, while UDA models align feature spaces between source and target domains, misalignment may happen in label spaces especially when majority classes are different across domains.
To reduce label distribution discrepancy, we propose an invariable prediction reward to jointly incorporate feature and label variants. 
Intuitively, we expect the classifier can make similar predictions by the original and masked features while reducing its dependence on domain-dependent patterns.
Therefore, our goal is to make a consistent prediction between $C(E(X))$ and $C(M(E(X)))$.
We follow the work~\cite{8578490} and employ L1-distance to measure the representation discrepancy loss between $C(E(X))$ and $C(M(E(X)))$ as the following:
\begin{equation}\label{rc}
    R_{c} = min(\mathcal{L}_{dis}(|C(M(E(X)))-C(E(X))|))
\end{equation}
, where $R_{c}$ measures cross-domain variations.
 
\subsection{Actor-Critic Learning}\label{step c}
To reduce the uncertainty of extracting domain-independent representations by the mask model, we adopt the policy gradient via the actor-critic algorithm~\cite{NIPS1999_6449f44a} to explore the optimal solution. 

First, we introduce a value estimation network, critic.
The critic helps to estimate an action's reward by giving a state.
Our critic is a 2-layer feed-forward network $f$ with the input of $M(E(X))$ and $E(X)$, the predictive reward $R_{p}$ is formulated as follow:
\begin{equation}
    R_{p} = f(M(E(X)) * E(X)).
\end{equation}
The loss function is the difference between of realistic reward ($R_{d}$ + $R_{c}$) and the predictive reward ($R_{p}$) from the critic as the following,
\begin{equation}
    \mathcal{L}(\theta_{c}) =\left(R_{d}+R_{c}-R_{p}\right)^{2}
\end{equation}
The critic is trained with Adam on a mean squared error $\mathcal{L}(\theta_{c})$. 

The mask model generates a mask matrix $\mathcal{M}_{a}$ and is an actor model by a fully connected neural network and a sigmoid unit.
It accepts inputs from the encoder and calculates a masked probability of each features ${\cal{M}}_{p}$.
Then we adopt Bernoulli sampling and obtain a logical matrix ${\cal{M}}_{a}$. The elements in ${\cal{M}}_{a}$ belongs to $\{0,1\}$.
We denote the output of the mask model as $x_{m} = \mathcal{M}_{a}* E(x)$.
The mask model's training objective is to maximum the total reward $R_{d}$ and $R_{c}$ defined in e.q.~\ref{rd} and e.q.~\ref{rc}
\begin{equation}\label{mask objective}
\begin{aligned}
    J({\cal{M}}_{a} \mid E(X))= \\
    \mathbb{E}_{{\cal{M}}_{a} \sim  \pi({\cal{M}}_{p} \mid E(X))} 
    \{R_{d} - R_{c} + R_{reg}\},
\end{aligned}
\end{equation}
, where $\pi$ is a policy function and $R_{reg}$ is a regularization term, controlling the number of masked features. We set $R_{reg} = (\sum {\cal{M}}_{a})$.
Since the mask model only make one action to transfer the state $M(E(X))$ from $E(X)$, we do not need to consider the future reward and the decay factor $\gamma$  in e.q.~\ref{op1} is zero.
Thus, we can obtain the following optimization by combining with the e.q.~\ref{op1} and e.q.~\ref{op2},
\begin{equation}
\begin{aligned}
    \mathcal{L}(\theta_{m}) = -\log \pi_{\theta_{m}}\left(a, s\right) A\left(s, a\right) + \\
    \pi_{\theta_{m}}(a \mid s) \log \pi_{\theta_{m}}(a \mid s)
\end{aligned}
\end{equation}
, where $A(s, a) = R_{d} +  R_{c} - R_{p}$. 
We update $\theta_{m}$ by minimizing $\mathcal{L}(\theta_{m})$.

\begin{algorithm}[ht!] 
\caption{Optimization Process of Our Model.}
\label{alg:Framwork} 
\begin{algorithmic}[1]
\REQUIRE 
	The source data $D_{s} = (X_{s},Y_{s})$ and target data $D_{t} = (X_{t})$, maximum iteration $I$;\\
\ENSURE
	The network parameter $\theta_{e},  \theta_{cla}, \theta_{d}, \theta_{m}$, $\theta_{c}$;
	\FOR{$i=1$; $i<I$; $i++$}
    \STATE Samples a batch from $D_{s}$ and $D_{t}$;
    \STATE Update $\theta_{e}, \theta_{cla}$ via e.q.(\ref{step a});
    \STATE Update $\theta_{d}$ via e.q.(\ref{discriminator})
    \STATE Update $\theta_{m}$,  $\theta_{c}$ via section (\ref{step c})
    \ENDFOR
\RETURN $\theta_{e}, \theta_{cla}, \theta_{d}, \theta_{m}$, $\theta_{c}$; 
\end{algorithmic}
\end{algorithm}

\subsection{Training Procedure}\label{Training Procedure}
Our training procedure includes three steps: 1) \textbf{step A} trains the encoder and classifier as e.q.~\ref{step a}; 2) \textbf{step B} trains the discriminator by e.q.~\ref{discriminator}; 3) \textbf{step C} training the mask model by the reinforcement learning.
We summarize the optimization process in Algorithm~\ref{alg:Framwork}.


\section{Experiment}
\begin{table*}[ht]
\centering
\resizebox{0.98\textwidth}{!}{
\begin{tabular}{c||c|c|c|c|c|c}
Method & MeToo - Davidson & Davidson-MeToo & Book-Kitchen & Kitchen-Book & Yelp-IMDB & IMDB-Yelp \\\hline
\multicolumn{7}{c}{LSTM} \\ \hline
DANN & 45.00 & 23.17 & 83.33 & 93.55 & 45.16 & 61.79 \\ \hline
MCD & 40.25 & 23.61 & 83.85 & 94.17 & 48.27 & 61.54 \\ \hline
JUMBOT & 46.94 & 23.26 & 81.79 & 93.66 & 42.57 & 56.78 \\ \hline
ALDA & 38.20 & 23.31 & 84.14 & 93.88 & 42.30 & 52.46 \\ \hline
URAM & 47.06 & 24.00 & 85.09 & 94.49 & 50.58 & 62.50 \\ \hline
\multicolumn{7}{c}{BERT} \\ \hline
DANN & 78.20 & 23.50 & 73.23 & 69.64 & 54.36 & 43.44 \\ \hline
MCD& 79.51 & 23.39 & 74.33 & 69.54 & 43.67 & 42.37 \\ \hline
JUMBOT & 73.74 & 23.23 & 80.57 & 75.00 & 53.37 & 43.08\\ \hline
ALDA & 77.26 & 24.42 & 77.21 & 70.54 & 47.01 & 39.84 \\ \hline
URAM & 81.93 & 27.09 & 86.24 & 76.97 & 57.70 & 45.16 \\
\end{tabular}}
\caption{Cross-domain performance of UDA models using F1 score. Each UDA model testifies over two popular neural feature extractor, LSTM and BERT. We list extensive evaluations in the Appendix.}
\label{overall}
\end{table*}

\subsection{Datasets}

\begin{table}[!ht]
\centering
\begin{tabular}{c|c|c|c}
 & Docs & Tokens & pos/neg \\\hline\hline
M-MeToo & 4480 & 13.86 & - \\
M-Davidson & 4480 & 19.13 & - \\\hline
A-Book & 2000 & 25.65  & 0.65 \\
A-Kitchen & 2000 & 29.73 & 4.78 \\\hline
Yelp & 2000 & 231.57 & 0.26 \\
IMDB & 2000 & 146.01 & 0.67\\
\end{tabular}
\caption{Data statistics summary of Morality and three review data, Amazon, Yelp and IMDB. We include multi-label distributions of the Morality data in appendix, Table~\ref{tab:moral_data}.}
\label{tab:data}
\end{table}

We assembled four datasets, three online reviews and one Twitter data.
The reviews are binary labels, and the Twitter data has 11 unique labels.
We summarize data statistics in Table~\ref{tab:data}.

\paragraph{Amazon, Yelp, and IMDB Review} are standard data sources for evaluating UDA models~\cite{ramponi2020neural}. 
We retrieved the \textbf{Yelp} and \textbf{IMDB} reviews\footnote{\url{https://pytorch.org/text/stable/datasets.html}} from torchtext and top four product genres of Amazon reviews~\cite{ni2019justifying}, including Books (B), DVDs (D), Electronics (E) and Kitchen (K).
We treat the four Amazon genres, Yelp, and IMDB as domains. 
Following the standard benchmark~\cite{blitzer-etal-2006-domain} for the UDA evaluations, we randomly select 2000 samples from each domain, while label distributions are not the same cross domains. 
We name cross-domain evaluations by the source-target format.
For example, Books-Kitchen means that Books is the source data and Kitchen is the target data.


\textbf{MFTC}~\cite{hoover2020moral} is a multi-label classification Twitter data with 35,108 tweets. 
These tweets are drawn from seven different discourse domains with moral sentiment across seven social movements, including MeToo, Black Lives Matter (BLM), Sandy, Davidson, Baltimore, All Lives Matter (ALM), and US Presidential Election (Election).
We treat social movements as domains.
These domains share the same set of 11 moral sentiment types: Subversion, Authority, Cheating, Fairness, Harm, Care, Betrayal, Loyalty, Purity, Degradation, Non-moral.
The rates of each of the virtues and vices vary substantially across the domain. 
For example, only approximately 2\% of
the ALM data were labeled as degradation while
approximately 14\% of the Sandy data were labeled as degradation.

\begin{table*}[htp]
\caption{Summary of domain shifts in data (domain-wise) and label (category-wise) distributions. }
\label{tab:domain_var}
\resizebox{\textwidth}{!}{
    \begin{tabular}{lcccccc}
    \hline
    discrepancy & MeToo-Davidson & Davidson-MeToo & Book-Kitchen & Kitchen-Book & Yelp-IMDB & IMDB-Yelp\\
    \hline
    domain-wise &10889  &661  &15986  &11680  &1.523  &1.692  \\
    \hline
    category-wise & 0.1197 & 0.1933 & $2.0 \times 10^{-4}$ & $1.0 \times 10^{-4}$ &0.044  & 0.050 \\
     \hline
    \end{tabular}
}
\end{table*}

We conduct an exploratory analysis of \textit{domain shifts} in data and labels.
The analysis follows the name format as source-target.
We use KL-divergence of the class distribution to measure the category-wise distribution and Euclidean distance to measure the domain-wise distribution. 
The domain-wise discrepancy refers to the euclidean distance of the encoder's output between the training and test sets. 
The category-wise is the KL-divergence of labels' distribution between the training and test sets.
We extract feature vectors using LSTMs trained over the domains. 
We show cross-domain discrepancy in Table~\ref{tab:domain_var}.
We can find that the multi-label Twitter data has more variations in both domain and label distributions.

\subsection{Baselines}
We compare our models with four recent methods.
\begin{itemize}
    \item DANN~\cite{ganin2015unsupervised}
    maps source and target domains to a common subspace through shared parameters. This approach introduces a gradient reversal layer to confuse domain prediction to improve classification robustness across domains with the adversarial train.
    \item MCD~\cite{Saito2018MaximumCD} proposes to maximize the discrepancy between two classifiers' outputs to detect target samples that are far from the support of the source. Then, A feature generator learns to generate target features near the support to minimize the discrepancy. 
    \item JUMBOT~\cite{DBLP:journals/corr/abs-2103-03606} proposes a new formulation of the mini-batch optimal transport strategy coupled with an unbalanced optimal transport program to calculate optimal transport distance.
    \item ALDA~\cite{chen2020adversarial} constructs a new loss function by introducing a confusion matrix.
    The confusion matrix reduces the gap and aligns the feature distributions in an adversarial manner.
\end{itemize}

\subsection{Implementation Details}

In this study, we evaluate the UDA methods using two standard neural models as feature extractors, LSTM~\cite{schmidhuber1997long} and BERT~\cite{devlin2019bert}. 
For the LSTM-based encoder, we use pre-trained word vectors GloVe~\cite{pennington2014glove} by torchtext~\footnote{https://pytorch.org/text/stable/index.html} to train word embedding.
The learning rate is set to $1 \times 10^{-3}$ and batch size set to 64.
We utilize a Bidirectional LSTM as our encoder and set the LSTM hidden number as $256$. 
For the BERT-based encoder, we load the pre-trained BERT model (\texttt{bert-base-uncased}) from the transformer toolkit~\cite{wolf2020transformers}.
We set the learning rate as $1 \times 10^{-5}$ and batch size as 16.

In all the above experiments, we used Adam \cite{DBLP:journals/corr/KingmaB14} to optimize our model and maximum iteration set to 50 in all experiments.
We run each experiment five times and average F1 as the final performance.

\subsection{Result}

\begin{table}[htp]
\caption{The domain-wise discrepancy based on domain adaptation methods.}
\resizebox{0.48\textwidth}{!}{
\begin{tabular}{lccccc}
\hline
 & DANN & MCD & JUMBOT & ALDA & URAM \\
\hline
MeToo - Davidson & 3.937 & 5.806 & 0.072 & 7.902 & 0.401 \\
Davidson-MeToo & 0.016 & 10.862 & 0.121 & 0.016 & 0.044 \\
Book-Kitchen & 0.950 & 1.651 & 0.046 & 3.922 & 0.233 \\
Kitchen-Book & 0.649 & 1.749 & 0.073 & 2.984 & 0.196 \\
Yelp-IMDB & 3.376 & 3.029 & 0.492 & 8.106 & 0.586 \\
IMDB-Yelp & 2.951 & 6.184 & 0.733 & 31.469 & 0.665 \\
\hline
\end{tabular}}
\end{table}

In this section, we present model performance on the cross-domain adaptation task and conduct an ablation analysis to examine the effects of the two reward factors, $R_{d}$ and $R_{c}$. We include extensive evaluation results in the appendix (Section~\ref{appendix:eval}).

\textbf{Overall Performance}. The table~\ref{overall} reports the overall performance.
Our method achieves the best result in the datasets with a significant discrepancy both in domain and category. 
We obtain a significant improvement on Amazon datasets, Book-Kitchen (1.12\%-17.7\%) and Kitchen-Book (2.62\%-10.68\%), respectively.
Amazon datasets follow the traditional assumption that different domains have significant feature discrepancies but have similar label distributions.
Our improvement on Amazon datasets verifies our model effectiveness of learning transferable knowledge.
On the other hand, our method also can release the category discrepancy problem.
As shown in the table~\ref{overall}, our method outperforms existing methods remarkably on the MFTC dataset (Metoo-Davidson) with the significant discrepancy in domain and category since we can align the distribution both in-text features and labels.
We notice some latest methods fail to compete with DANN.
We infer the reasons behind this are that some methods do not consider category discrepancy. 
For example, the performance of ALDA is lower than DANN on Metoo-Davidson since ALDA tries to align category discrepancy by narrowing domain discrepancy, which causes negative knowledge transfer.
The other reason is due to poor robustness. 
Some methods may ascribe samples' feature discrepancy to domain discrepancy, and aligning these sample's specific features lead to a lower distinguished ability among different samples (e.g., ALDA on Yelp-IMDB). 
All methods have similar performance on Davidson-Metoo since Davidson datasets have an extreme label distribution.
Most samples focus on the same category, which causes models not to access enough samples to learn the features in other classes.

\begin{figure}[htp]
\centering 
\includegraphics[width=0.48\textwidth]{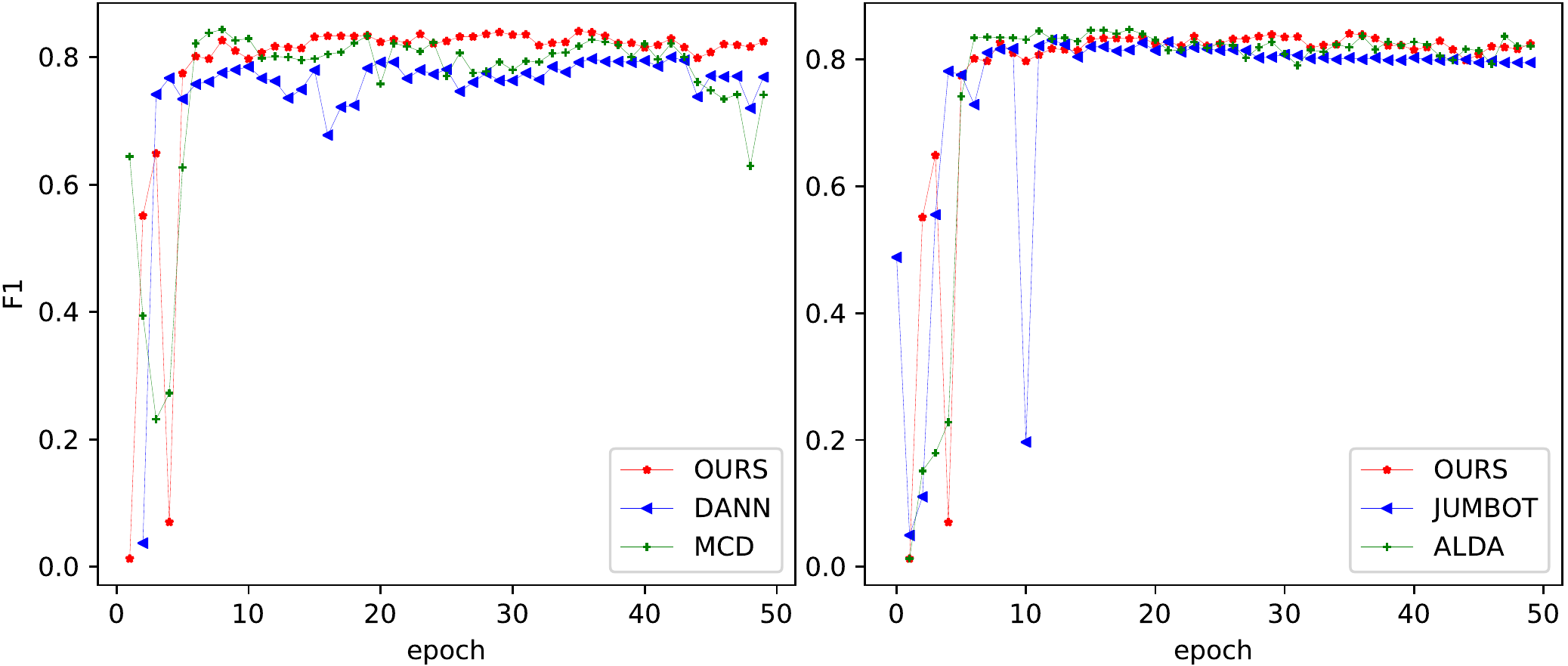}
\caption{The convergence comparison between our model and baselines on Book-Kitchen.}
\label{convergence}
\end{figure}

\textbf{Convergence Investigation} The convergence curves of our model and baselines are respectively depicted in Fig.~(\ref{convergence}).
We conduct a convergence experiment on Book-Kitchen datasets based on LSTM to verify the training stability during knowledge transfer. 
This task focuses on evaluating the ability to align domain-wise discrepancy since the feature's center of Book and Kitchen have a remarkable difference (up to 15986), but their categories are similar.
Specifically, we observe that our model significantly outperforms DANN and MCD during training.
DANN has relatively low stability since it only aligns different domain features without considering task-specific features.
Compared with ALDA, our model achieves similar stability.
Our model can achieve efficient convergence after iterating 15 epochs, which proves our model's robustness.

\begin{table*}[ht!]
\caption{Ablation studies of our model on LSTM}
\resizebox{0.98\textwidth}{!}{
\begin{tabular}{ccccccc}
\hline
Method & MeToo - Davidson & Davidson-MeToo & Book-Kitchen & Kitchen-Book & Yelp-IMDB & IMDB-Yelp \\
\hline
$-R_{d}$ & 31.97 & 23.14 & 86.09 & 84.15 & 43.28 & 61.40 \\
$-R_{c}$ & 35.84 & 23.19 & 84.51 & 80.87 & 43.71 & 60.87 \\
URAM &47.06 &24.00 &85.09  &94.49 &50.58  &62.50\\
\hline
\label{tab:ablation}
\end{tabular}
}
\end{table*}

\textbf{Knowledge Transfer.} We measure the feature center distance between the training set in the source data and the test set in the target data to evaluate models' ability to transfer knowledge.
Generally, the domain-wise discrepancy is significantly narrowed after applying domain adaptation methods. 
Our model achieves relatively significant improvements, but there are some exceptions.
For example, ALDA has a lower domain-wise discrepancy on Davidson-MeToo than ours. 
However, ALDA's performance is unsatisfactory, especially when the datasets have similar domains (e.g., Yelp-IMDB and IMDB-Yelp).
A similar situation also happens on DANN and MCD.
These methods enlarge domain-wise discrepancy when the domains have similar feature distribution.
Compared with JUMBOT, our model has a slightly large domain-wise discrepancy.
However, our model is more efficient on knowledge transfer when the domain has huge category-wise discrepancies. For example, the distance of our model is .0438 on Davidson-MeToo, while the corresponding figure is .1207 on JUMBOT.

\subsection{Ablation Analysis}

In this subsection, we investigate the importance of different rewards in reinforcement learning by conducting variant experiments, as shown in the Table~\ref{tab:ablation}. 

$-R_{c}$ means we delete reward $R_{c}$ in our $R_{adv}$. 
$R_{c}$ is a unsupervised reward.
Instead of aligning features, $R_{c}$ aims to search subspace features,  ensuing the consistent prediction between completed features $E(X)$ and sub-spaced features $M(E(X))$.
This method is efficient since removing $R_{c}$ is significantly detrimental to cross-domain performance.
Especially, we find that $R_{c}$ plays a more critical role Book-Kitchen and Kitchen-Book tasks by comparing the $R_{d}$ since removing $R_{c}$ lower the performance than $R_{d}$.

$R_{d}$ is proposed to align domain features by fooling the discriminator. 
$-R_{d}$ means we do not need to train the discriminator and $R_{adv}$ only combines with $R_{c}$ and $R_{reg}$.
$-R_{d}$ achieves a better performance than our completed model on Book-Kitchen.
We infer the reason behind this is because $R_{d}$ only focuses on feature shift rather than considering the discrepancy among different classes, which causes class-specific features to be weakened, and the model fails to distinguish the boundaries of other classes.
However, removing $R_{d}$ decreases the performance in most of the situations, which proves feature shift is efficient in domain adaptation.

Generally, $R_{d}$ and $R_{c}$ work together to guide critical knowledge transfer and removing any one of them significantly degrades the performance. 
Which reward dominates an improvement depends on the datasets' property.
When the domains have significant discrepancy both in features and label distribution, $R_{d}$ and $R_{c}$ work in an adversarial way to ensure shifting features as well as keeping class-specific features.


\section{Related work}

\paragraph{Unsupervised Domain Adaptation} for text classification has several major types of approaches, pivot features~\cite{blitzer-etal-2006-domain, daume2007frustratingly, ziser-reichart-2018-pivot, ben-david2020perl}, instance weighting~\cite{jiang2007instance, wang2019better, gong2020unified}, and domain adversaries~\cite{ganin2015unsupervised, qu-etal-2019-adversarial, du-etal-2020-adversarial}.
A recent survey~\cite{ramponi2020neural} shows that the most widespread methods for neural UDA are based on the use of domain adversaries, which reduces the discrepancy between the source and target distributions by reversing gradient updates for domain prediction networks.
Our study follows the same track to obtain domain-invariant representations, however, there are two major differences than the existing UDA for text classifiers: 1) a mask model to distill domain-invariant features, 2) and a reinforcement learning approach to optimize the adversarial network.
While existing UDA models have not explicitly incorporate domain shifts in label distributions, our proposed URAM jointly models domain variants in both data and label shifts.


\paragraph{Reinforcement Learning} 
With the robustness in learning sophisticated policies, recent works introduce Reinforcement learning (RL) into the unsupervised domain adaptation task~\cite{chen2020domain,10.1007/978-3-030-58598-3_44f,9423334}.
DARL~\cite{chen2020domain} employs deep Q-learning in partial domain adaptation.
The DARL framework designs a reward for the agent-based on how relevant the selected source instances are to the target domain. 
With the action-value function optimizer, DARL can automatically select source instances in the shared classes for circumventing negative transfer as well as to simultaneously learn transferable features between domains by reducing the domain shift. However, DARL does not generalize to unsupervised domain adaptation. Highly relying on the rich labels in the source domain will cause failure when insufficient labels are in the target domain.
To address this problem, \cite{9423334} develop a new reward across both source and target domains.
This reward can guide the agent to learn the best policy and select the closest feature pair for both domains.
However, rarely study has deployed the reinforcement UDA into the class-imbalanced text classification. 
To our best knowledge, we are the first work introducing RL for the UDA under the class-imbalanced text classification.

\paragraph{Imbalanced-class} 
Increasing works study the class-imbalanced domain adaptation~\cite{10.1007/978-3-030-66415-2_38,DBLP:journals/corr/abs-2012-12545, bose2021unsupervised, li2020dice}.
COAL~\cite{10.1007/978-3-030-66415-2_38} deals with feature shift and label shift in a unified way.
With the idea of prototype-based conditional distribution alignment and class-balanced self-training, COAL tackles feature shift in the context of label shift.
However, present works only focus on computer vision, and the imbalanced class domain adaptation in NLP is unexplored.
The other similar works is category-level feature alignment~\cite{qu-etal-2019-adversarial, 8954024, Li2021CategoryDG, 8851925, yang2020bi}.
These works usually focus on domain shifts and propose domain-level aligned strategies while ignoring the local category-level distributions, reducing cross-domain text classifiers' effectiveness.
A popular strategy for category-level alignment is aligning the same class features among different domains respectively by resorting to pseudo labels~\cite{10.1007/978-3-030-58598-3_44f,yang2020bi}.

\section{Conclusion}
In this study, we have proposed an unsupervised reinforcement adaptation model (URAM) for the novel cross-domain adaptation challenge where the source and target domains are class-imbalanced.
We demonstrate the effectiveness of our reinforcement approach with the other four state-of-art baselines on the task of text classification.
The URAM learns domain-independent representations by leveraging three reward factors, label, domain, and domain distance, which coherently combines pivot and adversarial approaches in UDA.
Extensive experiments and ablation analysis show that the URAM can obtain robust domain-invariant representations and effectively adapt text classifiers over both domains and imbalanced data.

\subsection{Limitation and Future Work}
Our work opens several future directions on the limitations of this study. 
First, class-imbalanced data naturally exist in NLP tasks, such as discourse inference~\cite{spangher2021multitask}, text generation~\cite{nishino2020reinforcement}, and question answering~\cite{li2020dice}.
Our next step will examine the effectiveness of our model over the NLP tasks.
Second, we only validate the URAM on English datasets, and additional multilingual settings will be verified in future work, such as multilingual text classification~\cite{schwenk2018corpus}.

\section{Acknowledgement}
The authors thank reviewers for their valuable comments. The work is partially supported by an internal award from the University of Memphis.

\bibliography{acl_latex}

\begin{thebibliography}{40}
\expandafter\ifx\csname natexlab\endcsname\relax\def\natexlab#1{#1}\fi

\bibitem[{Ben-David et~al.(2020)Ben-David, Rabinovitz, and
  Reichart}]{ben-david2020perl}
Eyal Ben-David, Carmel Rabinovitz, and Roi Reichart. 2020.
\newblock \href {https://doi.org/10.1162/tacl_a_00328} {{PERL}: Pivot-based
  domain adaptation for pre-trained deep contextualized embedding models}.
\newblock \emph{Transactions of the Association for Computational Linguistics},
  8:504--521.

\bibitem[{Blitzer et~al.(2006)Blitzer, McDonald, and
  Pereira}]{blitzer-etal-2006-domain}
John Blitzer, Ryan McDonald, and Fernando Pereira. 2006.
\newblock \href {https://aclanthology.org/W06-1615} {Domain adaptation with
  structural correspondence learning}.
\newblock In \emph{Proceedings of the 2006 Conference on Empirical Methods in
  Natural Language Processing}, pages 120--128, Sydney, Australia. Association
  for Computational Linguistics.

\bibitem[{Bose et~al.(2021)Bose, Illina, and Fohr}]{bose2021unsupervised}
Tulika Bose, Irina Illina, and Dominique Fohr. 2021.
\newblock \href {https://doi.org/10.18653/v1/2021.socialnlp-1.10} {Unsupervised
  domain adaptation in cross-corpora abusive language detection}.
\newblock In \emph{Proceedings of the Ninth International Workshop on Natural
  Language Processing for Social Media}, pages 113--122, Online. Association
  for Computational Linguistics.

\bibitem[{Chen et~al.(2020{\natexlab{a}})Chen, Wu, Duan, and
  Gao}]{chen2020domain}
Jin Chen, Xinxiao Wu, Lixin Duan, and Shenghua Gao. 2020{\natexlab{a}}.
\newblock \href {https://doi.org/10.1109/TNNLS.2020.3028078} {Domain
  adversarial reinforcement learning for partial domain adaptation}.
\newblock \emph{IEEE Transactions on Neural Networks and Learning Systems},
  pages 1--15.

\bibitem[{Chen et~al.(2020{\natexlab{b}})Chen, Zhao, Liu, and
  Cai}]{chen2020adversarial}
Minghao Chen, Shuai Zhao, Haifeng Liu, and Deng Cai. 2020{\natexlab{b}}.
\newblock Adversarial-learned loss for domain adaptation.
\newblock In \emph{Proceedings of the AAAI Conference on Artificial
  Intelligence}, volume~34, pages 3521--3528.

\bibitem[{Cheng et~al.(2020)Cheng, Guo, Candan, and
  Liu}]{cheng2020representation}
Lu~Cheng, Ruocheng Guo, K~Sel{\c{c}}uk Candan, and Huan Liu. 2020.
\newblock Representation learning for imbalanced cross-domain classification.
\newblock In \emph{Proceedings of the 2020 SIAM international conference on
  data mining}, pages 478--486. SIAM.

\bibitem[{Cui et~al.(2017)Cui, Coenen, and Bollegala}]{cui2017effect}
Xia Cui, Frans Coenen, and Danushka Bollegala. 2017.
\newblock Effect of data imbalance on unsupervised domain adaptation of
  part-of-speech tagging and pivot selection strategies.
\newblock In \emph{First International Workshop on Learning with Imbalanced
  Domains: Theory and Applications}, pages 103--115. PMLR.

\bibitem[{Daum\'{e}~III(2007)}]{daume2007frustratingly}
Hal Daum\'{e}~III. 2007.
\newblock \href {https://www.aclweb.org/anthology/papers/P/P07/P07-1033/}
  {Frustratingly easy domain adaptation}.
\newblock In \emph{Proceedings of the 45th Annual Meeting of the Association of
  Computational Linguistics}, pages 256--263.

\bibitem[{Devlin et~al.(2019)Devlin, Chang, Lee, and
  Toutanova}]{devlin2019bert}
Jacob Devlin, Ming-Wei Chang, Kenton Lee, and Kristina Toutanova. 2019.
\newblock \href {https://doi.org/10.18653/v1/N19-1423} {{BERT}: Pre-training of
  deep bidirectional transformers for language understanding}.
\newblock In \emph{Proceedings of the 2019 Conference of the North {A}merican
  Chapter of the Association for Computational Linguistics: Human Language
  Technologies, Volume 1 (Long and Short Papers)}, pages 4171--4186,
  Minneapolis, Minnesota. Association for Computational Linguistics.

\bibitem[{Dong et~al.(2020)Dong, Cong, Sun, Liu, and
  Xu}]{10.1007/978-3-030-58598-3_44f}
Jiahua Dong, Yang Cong, Gan Sun, Yuyang Liu, and Xiaowei Xu. 2020.
\newblock Cscl: Critical semantic-consistent learning for unsupervised domain
  adaptation.
\newblock In \emph{Computer Vision -- ECCV 2020}, pages 745--762, Cham.
  Springer International Publishing.

\bibitem[{Du et~al.(2020)Du, Sun, Wang, Qi, and
  Liao}]{du-etal-2020-adversarial}
Chunning Du, Haifeng Sun, Jingyu Wang, Qi~Qi, and Jianxin Liao. 2020.
\newblock \href {https://doi.org/10.18653/v1/2020.acl-main.370} {Adversarial
  and domain-aware {BERT} for cross-domain sentiment analysis}.
\newblock In \emph{Proceedings of the 58th Annual Meeting of the Association
  for Computational Linguistics}, pages 4019--4028, Online. Association for
  Computational Linguistics.

\bibitem[{Fatras et~al.(2021)Fatras, S{\'{e}}journ{\'{e}}, Courty, and
  Flamary}]{DBLP:journals/corr/abs-2103-03606}
Kilian Fatras, Thibault S{\'{e}}journ{\'{e}}, Nicolas Courty, and R{\'{e}}mi
  Flamary. 2021.
\newblock \href {http://arxiv.org/abs/2103.03606} {Unbalanced minibatch optimal
  transport; applications to domain adaptation}.
\newblock \emph{CoRR}, abs/2103.03606.

\bibitem[{Ganin and Lempitsky(2015)}]{ganin2015unsupervised}
Yaroslav Ganin and Victor Lempitsky. 2015.
\newblock Unsupervised domain adaptation by backpropagation.
\newblock In \emph{Proceedings of the 32nd International Conference on
  International Conference on Machine Learning - Volume 37}, ICML'15, page
  1180–1189. JMLR.org.

\bibitem[{Gong et~al.(2020)Gong, Yu, and Xia}]{gong2020unified}
Chenggong Gong, Jianfei Yu, and Rui Xia. 2020.
\newblock \href {https://doi.org/10.18653/v1/2020.emnlp-main.572} {Unified
  feature and instance based domain adaptation for aspect-based sentiment
  analysis}.
\newblock In \emph{Proceedings of the 2020 Conference on Empirical Methods in
  Natural Language Processing (EMNLP)}, pages 7035--7045, Online. Association
  for Computational Linguistics.

\bibitem[{Hochreiter and Schmidhuber(1997)}]{schmidhuber1997long}
Sepp Hochreiter and J\"{u}rgen Schmidhuber. 1997.
\newblock \href {https://doi.org/10.1162/neco.1997.9.8.1735} {Long short-term
  memory}.
\newblock \emph{Neural Comput.}, 9(8):1735–1780.

\bibitem[{Hoover et~al.(2020)Hoover, Portillo-Wightman, Yeh, Havaldar, Davani,
  Lin, Kennedy, Atari, Kamel, Mendlen, Moreno, Park, Chang, Chin, Leong, Leung,
  Mirinjian, and Dehghani}]{hoover2020moral}
Joe Hoover, Gwenyth Portillo-Wightman, Leigh Yeh, Shreya Havaldar,
  Aida~Mostafazadeh Davani, Ying Lin, Brendan Kennedy, Mohammad Atari, Zahra
  Kamel, Madelyn Mendlen, Gabriela Moreno, Christina Park, Tingyee~E. Chang,
  Jenna Chin, Christian Leong, Jun~Yen Leung, Arineh Mirinjian, and Morteza
  Dehghani. 2020.
\newblock \href {https://doi.org/10.1177/1948550619876629} {Moral foundations
  twitter corpus: A collection of 35k tweets annotated for moral sentiment}.
\newblock \emph{Social Psychological and Personality Science},
  11(8):1057--1071.

\bibitem[{Jiang and Zhai(2007)}]{jiang2007instance}
Jing Jiang and ChengXiang Zhai. 2007.
\newblock \href {https://aclanthology.org/P07-1034} {Instance weighting for
  domain adaptation in {NLP}}.
\newblock In \emph{Proceedings of the 45th Annual Meeting of the Association of
  Computational Linguistics}, pages 264--271, Prague, Czech Republic.
  Association for Computational Linguistics.

\bibitem[{Kingma and Ba(2015)}]{DBLP:journals/corr/KingmaB14}
Diederik~P. Kingma and Jimmy Ba. 2015.
\newblock \href {http://arxiv.org/abs/1412.6980} {Adam: {A} method for
  stochastic optimization}.
\newblock In \emph{3rd International Conference on Learning Representations,
  {ICLR} 2015, San Diego, CA, USA, May 7-9, 2015, Conference Track
  Proceedings}.

\bibitem[{Konda and Tsitsiklis(2000)}]{NIPS1999_6449f44a}
Vijay Konda and John Tsitsiklis. 2000.
\newblock \href
  {https://proceedings.neurips.cc/paper/1999/file/6449f44a102fde848669bdd9eb6b76fa-Paper.pdf}
  {Actor-critic algorithms}.
\newblock In \emph{Advances in Neural Information Processing Systems},
  volume~12. MIT Press.

\bibitem[{Lee et~al.(2020)Lee, Hyun, Seong, and
  Kim}]{DBLP:journals/corr/abs-2012-12545}
Suhyeon Lee, Junhyuk Hyun, Hongje Seong, and Euntai Kim. 2020.
\newblock \href {http://arxiv.org/abs/2012.12545} {Unsupervised domain
  adaptation for semantic segmentation by content transfer}.
\newblock \emph{CoRR}, abs/2012.12545.

\bibitem[{Li et~al.(2019)Li, He, Li, and Yang}]{8851925}
Lusi Li, Haibo He, Jie Li, and Guang Yang. 2019.
\newblock \href {https://doi.org/10.1109/IJCNN.2019.8851925} {Adversarial
  domain adaptation via category transfer}.
\newblock In \emph{2019 International Joint Conference on Neural Networks
  (IJCNN)}, pages 1--8.

\bibitem[{Li et~al.(2021)Li, Huang, Hua, and Zhang}]{Li2021CategoryDG}
Shuai Li, Jianqiang Huang, Xiansheng Hua, and Lei Zhang. 2021.
\newblock Category dictionary guided unsupervised domain adaptation for object
  detection.
\newblock In \emph{AAAI}.

\bibitem[{Li et~al.(2020)Li, Sun, Meng, Liang, Wu, and Li}]{li2020dice}
Xiaoya Li, Xiaofei Sun, Yuxian Meng, Junjun Liang, Fei Wu, and Jiwei Li. 2020.
\newblock \href {https://doi.org/10.18653/v1/2020.acl-main.45} {Dice loss for
  data-imbalanced {NLP} tasks}.
\newblock In \emph{Proceedings of the 58th Annual Meeting of the Association
  for Computational Linguistics}, pages 465--476, Online. Association for
  Computational Linguistics.

\bibitem[{Long et~al.(2018)Long, CAO, Wang, and Jordan}]{long2018conditional}
Mingsheng Long, ZHANGJIE CAO, Jianmin Wang, and Michael~I Jordan. 2018.
\newblock \href
  {https://proceedings.neurips.cc/paper/2018/file/ab88b15733f543179858600245108dd8-Paper.pdf}
  {Conditional adversarial domain adaptation}.
\newblock In \emph{Advances in Neural Information Processing Systems},
  volume~31, page~11. Curran Associates, Inc.

\bibitem[{Luo et~al.(2019)Luo, Zheng, Guan, Yu, and Yang}]{8954024}
Yawei Luo, Liang Zheng, Tao Guan, Junqing Yu, and Yi~Yang. 2019.
\newblock \href {https://doi.org/10.1109/CVPR.2019.00261} {Taking a closer look
  at domain shift: Category-level adversaries for semantics consistent domain
  adaptation}.
\newblock In \emph{2019 IEEE/CVF Conference on Computer Vision and Pattern
  Recognition (CVPR)}, pages 2502--2511.

\bibitem[{Ni et~al.(2019)Ni, Li, and McAuley}]{ni2019justifying}
Jianmo Ni, Jiacheng Li, and Julian McAuley. 2019.
\newblock \href {https://doi.org/10.18653/v1/D19-1018} {Justifying
  recommendations using distantly-labeled reviews and fine-grained aspects}.
\newblock In \emph{Proceedings of the 2019 Conference on Empirical Methods in
  Natural Language Processing and the 9th International Joint Conference on
  Natural Language Processing (EMNLP-IJCNLP)}, pages 188--197, Hong Kong,
  China. Association for Computational Linguistics.

\bibitem[{Nishino et~al.(2020)Nishino, Ozaki, Momoki, Taniguchi, Kano, Nakano,
  Tagawa, Taniguchi, Ohkuma, and Nakamura}]{nishino2020reinforcement}
Toru Nishino, Ryota Ozaki, Yohei Momoki, Tomoki Taniguchi, Ryuji Kano, Norihisa
  Nakano, Yuki Tagawa, Motoki Taniguchi, Tomoko Ohkuma, and Keigo Nakamura.
  2020.
\newblock \href {https://doi.org/10.18653/v1/2020.findings-emnlp.202}
  {Reinforcement learning with imbalanced dataset for data-to-text medical
  report generation}.
\newblock In \emph{Findings of the Association for Computational Linguistics:
  EMNLP 2020}, pages 2223--2236, Online. Association for Computational
  Linguistics.

\bibitem[{Pennington et~al.(2014)Pennington, Socher, and
  Manning}]{pennington2014glove}
Jeffrey Pennington, Richard Socher, and Christopher~D. Manning. 2014.
\newblock \href {http://www.aclweb.org/anthology/D14-1162} {Glove: Global
  vectors for word representation}.
\newblock In \emph{Empirical Methods in Natural Language Processing (EMNLP)},
  pages 1532--1543.

\bibitem[{Qu et~al.(2019)Qu, Zou, Cheng, Yang, and
  Zhou}]{qu-etal-2019-adversarial}
Xiaoye Qu, Zhikang Zou, Yu~Cheng, Yang Yang, and Pan Zhou. 2019.
\newblock \href {https://doi.org/10.18653/v1/N19-1258} {Adversarial category
  alignment network for cross-domain sentiment classification}.
\newblock In \emph{Proceedings of the 2019 Conference of the North {A}merican
  Chapter of the Association for Computational Linguistics: Human Language
  Technologies, Volume 1 (Long and Short Papers)}, pages 2496--2508,
  Minneapolis, Minnesota. Association for Computational Linguistics.

\bibitem[{Ramponi and Plank(2020)}]{ramponi2020neural}
Alan Ramponi and Barbara Plank. 2020.
\newblock \href {https://doi.org/10.18653/v1/2020.coling-main.603} {Neural
  unsupervised domain adaptation in {NLP}{---}{A} survey}.
\newblock In \emph{Proceedings of the 28th International Conference on
  Computational Linguistics}, pages 6838--6855, Barcelona, Spain (Online).
  International Committee on Computational Linguistics.

\bibitem[{Saito et~al.(2018{\natexlab{a}})Saito, Watanabe, Ushiku, and
  Harada}]{Saito2018MaximumCD}
Kuniaki Saito, Kohei Watanabe, Y.~Ushiku, and Tatsuya Harada.
  2018{\natexlab{a}}.
\newblock Maximum classifier discrepancy for unsupervised domain adaptation.
\newblock \emph{2018 IEEE/CVF Conference on Computer Vision and Pattern
  Recognition}, pages 3723--3732.

\bibitem[{Saito et~al.(2018{\natexlab{b}})Saito, Watanabe, Ushiku, and
  Harada}]{8578490}
Kuniaki Saito, Kohei Watanabe, Yoshitaka Ushiku, and Tatsuya Harada.
  2018{\natexlab{b}}.
\newblock \href {https://doi.org/10.1109/CVPR.2018.00392} {Maximum classifier
  discrepancy for unsupervised domain adaptation}.
\newblock In \emph{2018 IEEE/CVF Conference on Computer Vision and Pattern
  Recognition}, pages 3723--3732.

\bibitem[{Schwenk and Li(2018)}]{schwenk2018corpus}
Holger Schwenk and Xian Li. 2018.
\newblock \href {https://aclanthology.org/L18-1560} {A corpus for multilingual
  document classification in eight languages}.
\newblock In \emph{Proceedings of the Eleventh International Conference on
  Language Resources and Evaluation ({LREC} 2018)}, Miyazaki, Japan. European
  Language Resources Association (ELRA).

\bibitem[{Spangher et~al.(2021)Spangher, May, Shiang, and
  Deng}]{spangher2021multitask}
Alexander Spangher, Jonathan May, Sz-Rung Shiang, and Lingjia Deng. 2021.
\newblock \href {https://doi.org/10.18653/v1/2021.emnlp-main.40} {Multitask
  semi-supervised learning for class-imbalanced discourse classification}.
\newblock In \emph{Proceedings of the 2021 Conference on Empirical Methods in
  Natural Language Processing}, pages 498--517, Online and Punta Cana,
  Dominican Republic. Association for Computational Linguistics.

\bibitem[{Tan et~al.(2020)Tan, Peng, and Saenko}]{10.1007/978-3-030-66415-2_38}
Shuhan Tan, Xingchao Peng, and Kate Saenko. 2020.
\newblock Class-imbalanced domain adaptation: An empirical odyssey.
\newblock In \emph{Computer Vision -- ECCV 2020 Workshops}, pages 585--602,
  Cham. Springer International Publishing.

\bibitem[{Wang et~al.(2019)Wang, Bi, Wang, and Liu}]{wang2019better}
Zhi Wang, Wei Bi, Yan Wang, and Xiaojiang Liu. 2019.
\newblock \href {https://doi.org/10.1609/aaai.v33i01.33017241} {Better
  fine-tuning via instance weighting for text classification}.
\newblock In \emph{Proceedings of the Thirty-Third AAAI Conference on
  Artificial Intelligence and Thirty-First Innovative Applications of
  Artificial Intelligence Conference and Ninth AAAI Symposium on Educational
  Advances in Artificial Intelligence}, AAAI'19/IAAI'19/EAAI'19. AAAI Press.

\bibitem[{Wolf et~al.(2020)Wolf, Debut, Sanh, Chaumond, Delangue, Moi, Cistac,
  Rault, Louf, Funtowicz, Davison, Shleifer, von Platen, Ma, Jernite, Plu, Xu,
  Le~Scao, Gugger, Drame, Lhoest, and Rush}]{wolf2020transformers}
Thomas Wolf, Lysandre Debut, Victor Sanh, Julien Chaumond, Clement Delangue,
  Anthony Moi, Pierric Cistac, Tim Rault, Remi Louf, Morgan Funtowicz, Joe
  Davison, Sam Shleifer, Patrick von Platen, Clara Ma, Yacine Jernite, Julien
  Plu, Canwen Xu, Teven Le~Scao, Sylvain Gugger, Mariama Drame, Quentin Lhoest,
  and Alexander Rush. 2020.
\newblock \href {https://doi.org/10.18653/v1/2020.emnlp-demos.6} {Transformers:
  State-of-the-art natural language processing}.
\newblock In \emph{Proceedings of the 2020 Conference on Empirical Methods in
  Natural Language Processing: System Demonstrations}, pages 38--45, Online.
  Association for Computational Linguistics.

\bibitem[{Yang et~al.(2020)Yang, Xia, Ding, and Ding}]{yang2020bi}
Guanglei Yang, Haifeng Xia, Mingli Ding, and Zhengming Ding. 2020.
\newblock Bi-directional generation for unsupervised domain adaptation.
\newblock In \emph{Proceedings of the AAAI Conference on Artificial
  Intelligence}, volume~34, pages 6615--6622.

\bibitem[{Zhang et~al.(2021)Zhang, Ye, and Davison}]{9423334}
Youshan Zhang, Hui Ye, and Brian~D. Davison. 2021.
\newblock \href {https://doi.org/10.1109/WACV48630.2021.00068} {Adversarial
  reinforcement learning for unsupervised domain adaptation}.
\newblock In \emph{2021 IEEE Winter Conference on Applications of Computer
  Vision (WACV)}, pages 635--644.

\bibitem[{Ziser and Reichart(2018)}]{ziser-reichart-2018-pivot}
Yftah Ziser and Roi Reichart. 2018.
\newblock \href {https://doi.org/10.18653/v1/N18-1112} {Pivot based language
  modeling for improved neural domain adaptation}.
\newblock In \emph{Proceedings of the 2018 Conference of the North {A}merican
  Chapter of the Association for Computational Linguistics: Human Language
  Technologies, Volume 1 (Long Papers)}, pages 1241--1251, New Orleans,
  Louisiana. Association for Computational Linguistics.

\end{thebibliography}

\appendix

\section{Additional Data Statistics}

In this section, we summarize additional data and label statistics in Table~\ref{tab:amazon} and \ref{tab:moral_data}.

\begin{table}[!ht]
\centering
\begin{tabular}{c|c|c|c}
 & Docs & Tokens & pos/neg \\\hline\hline
D-DVD & 2000 & 30.51 & 2.52 \\
E-Electronic & 2000 & 27.65 & 2.26 \\\hline
\end{tabular}
\caption{Stats of the Amazon review data. We present the average number of tokens and the imbalanced-class ratio.}
\label{tab:amazon}
\end{table}

\begin{table*}[!ht]
\centering
\resizebox{1\textwidth}{!}{
    \begin{tabular}{c|ccccccccccc}\hline
    dataset & Non-moral & Degradation & Harm & Fairness & Subversion & Care & Cheating & Purity & Betrayal & Authority & Loyalty \\
    \hline
    MeToo & 21.40 & 15,30 & 6.86 &6.30 &14.70 &3.40 & 11.00 &2.98 & 5.83 &6.93 &5.29 \\
    BLM &23.59 &4.23 &19.36 &8.58 &5.74 &5.93 &13.84 &2.76 & 2.71 & 5.40 &7.83 \\
    Sandy & 13.68 &1.94 &1.69 &3.82 &9.63 &21.30 &9.80 &1.45 & 3.12 & 9.46 &8.86 \\
    Davidson &92.13 & 1.34 & 2.76 & 0.08 & 0.14 & 0.18 & 1.24 & 0.10 & 0.82 & 0.40 & 0.82 \\
    Baltimore &54.93 & 0.55 & 4.86 & 2.60 & 5.34 & 3.26 & 9.38 & 0.69 & 11.18 & 0.40 & 6.83 \\
    ALM & 20.98 & 3.18 & 19.15 & 13.42 & 2.37 & 11.88 & 13.16 & 2.11 & 1.04 & 6.36 & 6.36 \\
    Election &47.70 & 2.13 & 9.09 & 8.66 & 2.55 & 6.15 & 9.59 & 6.32 & 1.98 & 2.61 & 3.20 \\
    \hline
    \end{tabular}
}
\caption{Label distributions of the multi-class morality dataset~\cite{hoover2020moral}}
\label{tab:moral_data}
\end{table*}

\section{Cross-domain Evaluations}
\label{appendix:eval}

\begin{table*}[htp]
\centering
    \resizebox{1\textwidth}{!}{
    \begin{tabular}{cccccccc}
    \hline
    \textbf{No-adapt} & MeToo & BLM & Sandy & Davidson & Baltimore & ALM & Election \\
    \hline
    MeToo & 47.16 & 18.09 & 6.28 & 35.61 & 29.58 & 14.61 & 16.95 \\
    BLM & 16.23 & 76.32 & 17.22 & 26.27 & 25.28 & 16.16 & 26.40 \\
    Sandy & 8.81 & 14.46 & 58.50 & 19.27 & 7.49 & 15.68 & 9.04 \\
    Davidson & 23.12 & 31.98 & 8.09 & 99.17 & 66.96 & 24.93 & 58.49 \\
    Baltimore & 23.32 & 32.42 & 10.07 & 99.17 & 66.54 & 25.00 & 59.09 \\
    ALM & 12.11 & 17.60 & 14.27 & 24.88 & 25.12 & 43.71 & 20.33 \\
    Election & 23.18 & 32.59 & 15.24 & 99.11 & 66.57 & 24.95 & 58.87 \\
    \hline
    \end{tabular}
    \quad
    \begin{tabular}{cccccccc}
    \hline
    \textbf{MCD} & MeToo & BLM & Sandy & Davidson & Baltimore & ALM & Election \\
    \hline
   MeToo & 48.14 & 25.86 & 13.77 & 40.25 & 38.86 & 22.81 & 32.41 \\
    BLM & 16.48 & 78.42 & 17.27 & 29.17 & 55.27 & 23.40 & 34.51 \\
    Sandy & 24.37 & 16.68 & 60.17 & 15.74 & 32.50 & 16.52 & 12.58 \\
    Davidson & 23.62 & 31.99 & 13.94 & 99.17 & 66.96 & 25.73 & 58.49 \\
    Baltimore & 23.12 & 32.44 & 14.80 & 99.17 & 66.21 & 24.93 & 59.09 \\
    ALM & 16.88 & 23.37 & 15.48 & 37.11 & 34.33 & 63.18 & 25.22 \\
    Election & 23.12 & 32.53 & 14.10 & 99.17 & 66.54 & 24.93 & 63.91 \\
    \hline
    \end{tabular}
    }
    \resizebox{1\textwidth}{!}{
    \begin{tabular}{cccccccc}
    \textbf{DANN} & MeToo & BLM & Sandy & Davidson & Baltimore & ALM & Election \\\hline
    MeToo & 40.03 & 17.98 & 9.74 & 45.00 & 20.65 & 13.69 & 24.30 \\
    BLM & 16.33 & 75.40 & 15.48 & 35.68 & 22.94 & 17.82 & 24.39 \\
    Sandy & 8.37 & 14.55 & 56.84 & 6.78 & 6.47 & 14.65 & 9.34 \\
    Davidson & 23.17 & 31.98 & 8.17 & 99.17 & 66.96 & 24.93 & 58.49 \\
    Baltimore & 23.17 & 32.42 & 9.82 & 99.17 & 66.24 & 24.95 & 59.03 \\
    ALM & 12.63 & 16.78 & 14.93 & 19.18 & 20.87 & 60.88 & 17.26 \\
    Election & 23.14 & 32.57 & 14.23 & 99.17 & 66.57 & 24.93 & 64.01 \\
    \hline
    \end{tabular}
    \quad
    \begin{tabular}{cccccccc}
    \textbf{JUMBOT} & MeToo & BLM & Sandy & Davidson & Baltimore & ALM & Election \\\hline
    MeToo & 43.12 & 28.32 & 10.47 & 46.94 & 42.33 & 21.08 & 36.11 \\
    BLM & 24.37 & 72.57 & 16.02 & 52.20 & 48.92 & 32.18 & 48.91 \\
    Sandy & 19.34 & 33.17 & 57.60 & 10.86 & 41.23 & 30.86 & 39.59 \\
    Davidson & 23.26 & 32.99 & 8.35 & 99.17 & 66.96 & 26.64 & 58.49 \\
    Baltimore & 23.48 & 32.66 & 12.16 & 99.17 & 66.18 & 25.03 & 59.09 \\
    ALM & 23.30 & 39.82 & 17.04 & 66.60 & 61.70 & 42.01 & 46.50 \\
    Election & 23.12 & 32.49 & 15.20 & 99.17 & 66.42 & 24.93 & 60.41 \\
    \hline
    \end{tabular}
    }
    \resizebox{1\textwidth}{!}{
    \begin{tabular}{cccccccc}
    \textbf{ALDA} & MeToo & BLM & Sandy & Davidson & Baltimore & ALM & Election \\
    \hline
    MeToo & 21.50 & 25.89 & 14.17 & 38.21 & 1.12 & 9.84 & 58.75 \\
    BLM & 14.82 & 56.82 & 13.97 & 51.90 & 39.98 & 16.53 & 23.39 \\
    Sandy & 23.36 & 14.23 & 34.84 & 33.81 & 6.01 & 22.06 & 28.03 \\
    Davidson & 23.31 & 31.99 & 26.59 & 99.17 & 66.96 & 32.31 & 58.49 \\
    Baltimore & 23.03 & 31.63 & 8.77 & 42.12 & 65.33 & 25.50 & 28.77 \\
    ALM & 22.43 & 14.83 & 5.94 & 31.16 & 58.96 & 38.50 & 37.35 \\
    Election & 25.44 & 39.70 & 19.16 & 98.32 & 66.54 & 23.17 & 58.87 \\
    \hline
    \end{tabular}
    \quad
    \begin{tabular}{cccccccc}
    \textbf{URAM} & MeToo & BLM & Sandy & Davidson & Baltimore & ALM & Election \\
    \hline
    MeToo & 45.54 & 19.34 & 10.48 & 47.07 & 38.14 & 16.97 & 34.80 \\
    BLM & 16.03 & 79.12 & 15.86 & 50.31 & 30.57 & 18.56 & 26.74 \\
    Sandy & 9.28 & 14.65 & 60.44 & 10.50 & 10.28 & 15.28 & 8.86 \\
    Davidson & 24.00 & 32.53 & 11.59 & 99.17 & 66.96 & 25.02 & 58.49 \\
    Baltimore & 23.10 & 28.57 & 12.09 & 98.96 & 63.52 & 24.93 & 53.43 \\
    ALM & 12.58 & 16.51 & 15.70 & 34.43 & 27.88 & 63.11 & 17.29 \\
    Election & 22.54 & 31.92 & 12.38 & 99.06 & 58.10 & 24.88 & 65.23 \\
    \hline
    \end{tabular}
    }
\caption{Cross-domain performance evaluation over the Morality dataset~\cite{hoover2020moral} using F1. Each subtable presents results of one UDA model.}
\label{tab:moral_results_no}
\end{table*}
\begin{table*}
\centering
\resizebox{1\textwidth}{!}{
    \begin{tabular}{ccccccccccc}
    \hline
     & book-dvd & dvd-book & book-eletronic & eletronic-book & kitchen-eletronic & eletronic-kitchen & dvd-kitchen & kitchen-dvd & dvd-eletroic & eletronic-dvd \\
     \hline
    DANN & 83.16 & 94.00 & 86.87 & 92.15 & 95.24 & 91.21 & 94.24 & 94.29 & 94.63 & 92.57 \\
    MCD & 84.39 & 94.34 & 85.06 & 93.36 & 94.08 & 91.61 & 94.14 & 94.99 & 94.22 & 92.54 \\
    JUMBOT & 82.27 & 91.51 & 77.34 & 84.83 & 92.91 & 85.58 & 92.49 & 94.01 & 91.64 & 92.23 \\
    ALDA & 84.49 & 93.52 & 84.14 & 94.49 & 93.93 & 92.39 & 92.70 & 94.21 & 94.00 & 90.91 \\
    URAM & 86.56 & 94.58 & 87.90 & 93.51 & 94.96 & 92.87 & 94.81 & 95.15 & 95.03 & 93.02 \\
    \hline
    \end{tabular}
}
\caption{Cross-domain performance evaluation over the Amazon review dataset~\cite{blitzer-etal-2006-domain}.}
\label{tab:eval_amazon}
\end{table*}
\end{document}